**Applying Association Rules Mining to Investigate Pedestrian Fatal and Injury Crash Patterns Under Different Lighting Conditions**

**Ahmed Hossain**[*]
*(Corresponding Author)*
Ph.D. Student
Department of Civil Engineering
University of Louisiana at Lafayette, Lafayette, Louisiana, 70504
Email: ahmed.hossain1@louisiana.edu

**Xiaoduan Sun, Ph.D., P.E.**
Professor
Department of Civil Engineering
University of Louisiana at Lafayette, Lafayette, Louisiana, 70504
Email: xsun@louisiana.edu

**Raju Thapa, Ph.D., P.E.**
Assistant Professor, Research
Louisiana Transportation Research Center
Baton Rouge, Louisiana, 70808
Email: raju.thapa@la.gov

**Julius Codjoe, Ph.D., P.E.**
Assistant Professor, Research
Louisiana Transportation Research Center
Baton Rouge, Louisiana, 70808
Email: julius.codjoe@la.gov

*[Published on February 18, 2022]*

## ABSTRACT

The pattern of pedestrian crashes varies greatly depending on lighting circumstances, emphasizing the need of examining pedestrian crashes in various lighting conditions. Using Louisiana pedestrian fatal and injury crash data (2010-2019), this study applied Association Rules Mining (ARM) to identify the hidden pattern of crash risk factors according to three different lighting conditions (daylight, dark-with-streetlight, and dark-no-streetlight). Based on the generated rules, the results show that daylight pedestrian crashes are associated with children (<15 years), senior pedestrians (>64 years), older drivers (>64 years), and other driving behaviors such as 'failure to yield', 'inattentive/distracted', 'illness/fatigue/asleep'. Additionally, young drivers (15-24 years) are involved in severe pedestrian crashes in daylight conditions. This study also found pedestrian alcohol/drug involvement as the most frequent item in the dark-with-streetlight condition. This crash type is particularly associated with pedestrian action (crossing intersection/midblock), driver age (55-64 years), speed limit (30-35 mph), and specific area type (business with mixed residential area). Fatal pedestrian crashes are found to be associated with roadways with high-speed limits (>50 mph) during the dark without streetlight condition. Some other risk factors linked with 'high-speed limit' related crashes are pedestrians walking with/against the traffic, presence of pedestrian dark clothing, pedestrian alcohol/drug involvement. The research findings are expected to provide an improved understanding of the underlying relationships between pedestrian crash risk factors and specific lighting conditions. Highway safety experts can utilize these findings to conduct a decision-making process for selecting effective countermeasures to reduce pedestrian crashes strategically.

**INTRODUCTION**

Pedestrian crashes are a major traffic safety concern because they have a distinct pattern of high injury severity and mortality rate. According to the Fatality Analysis Reporting System (FARS) database, a total of 6,205 pedestrians were killed in traffic crashes accounting for 17% of all traffic fatalities in the United States in 2019 (1). The number of pedestrians killed in 2019 was at its highest level since 1990 (2). The situation is even far worse in Louisiana, which experiences a substantially higher pedestrian fatality rate (2.54 per 100,000 population) compared to the national average (1.89 per 100,000 population) (3). The AASHTO Strategic Highway Safety Plan prioritizes improved pedestrian safety as one of its top priorities (4).

Consider a hypothetical scenario in which a pedestrian (dressed in black cloth) crosses at a midblock location during daylight. From a visibility standpoint, this may not pose a serious safety risk because the driver might assess the situation and adjust their movement accordingly. However, the same setup could be lethal due to low visibility in the dark. Hence, the outcome of pedestrian crashes differs dramatically depending on the lighting conditions, which requires further investigation. Some previous studies identified lighting condition as one of the significant factor leading to pedestrian crashes (5–8). Most of this research tried to link lighting condition with pedestrian injury severity considering several risk factors such as crash location, pedestrian or driver age/gender, roadway geometry, and environmental characteristics. Additionally, the majority of these investigations made a general conclusion that fatal pedestrian crashes are more likely to occur at night whereas nonfatal pedestrian crashes are more likely to occur during the day (9). None of the prior studies looked at how the association of pedestrian crash contributing factors varies according to different lighting conditions.

Each crash is the result of a chain of critical events. Due to the dynamic and complex nature of an individual crash, different components such as humans, vehicles, roadways, and environmental factors actively interact with each other to cause a crash (10). Gaining this knowledge of interaction and association is the key to pedestrian safety. Data mining methods such as Association Rules Mining (ARM) can provide valuable insights by detecting crash patterns from a given database (11). By conducting ARM, this study investigates the unrevealed pattern of key crash attributes under different lighting conditions utilizing ten (2010 to 2019) years of pedestrian fatal and injury crashes in Louisiana state as a case study. Unlike most of the previous studies, which considered lighting condition as a single contributing factor, this research categorizes lighting conditions into three groups (daylight, dark-with-streetlight, and dark-no-streetlight) to identify the pattern of pedestrian crashes.

**Pedestrian Crash Risk Factors Under Different Lighting Conditions**

According to the literature, pedestrian crashes and their severity are influenced by several crash variables (12). Some of these variables are related to pedestrian demographics (age, gender, alcohol use, actions, primary contributing factors), driver demographics (age, gender, condition, violation type), roadway features (number of lanes, posted speed limits, intersection presence, functional classification) and other factors such as land use pattern, day of the week, weather and lighting condition (13–16). Ferguson et al. reported that poor lighting conditions combined with alcohol intake by pedestrians and high-speed limits on limited access routes were found to greatly exacerbate the risk of pedestrian injury severity (17). However, another study performed by Sullivan did not find any similar effect of alcohol intake among drivers involved in pedestrian crashes in the dark (18). The observed differences in pedestrian crash risk linked with drinking (whether it is drunk pedestrians or drunk drivers) may be interrelated to the crash exposure - intoxicated pedestrians engage in more risky road behavior compared to drunk drivers.

Very few of the previous studies have specifically focused on the relationship between lighting conditions and pedestrian crash risk factors. Seyedmirsajad Mokhtarimousavi utilized logistic regression and support vector machine to identify the contributing factors of pedestrian crash injury severity by daytime and nighttime in California (19). The author explored that some contributing factors were significant at daytime injury severity (parked motor vehicle in crash, rural freeways, roadway type, truck at fault, weather condition, and driver sobriety) while other factors were significant at nighttime injury severity (head-on crash, rainy weather condition, intersection location, pedestrian crossing in crosswalk, multilane undivided non-freeway roadway type). In another study conducted in California, Siddiqui et al. employed an ordered probit model to explore the impact of various risk factors on pedestrian injury severity (20). They focused explicitly on two types of crossing locations (midblock and intersection) and three types of lighting conditions (daylight, dark with streetlight, dark without streetlight). They concluded that severe pedestrian accidents occur more frequently at midblock than at intersections, and streetlights reduce fatal pedestrian crashes at both midblock and intersection locations. Although both of these studies evaluated the impact of lighting conditions on pedestrian crashes, none of them explored the association of contributing factors.

Nonetheless, a few recent research have used ARM to find major associations of contributory factors leading to pedestrian crashes (21–23). However, none of them focused on the relationship between lighting conditions and pedestrian crashes. Some other researchers utilized various regression models for pedestrian crash analysis such as binary logistic regression model (24), multinomial logistic regression model (25), partial proportional odds logit model (26), mixed logit model (27,28), ordered logit/probit model (29–31) and log-linear model (32). These studies are based on the assumption that crash contributing factors are independent of one another, which could lead to an erroneous assessment of every single factor (33). Considering this viewpoint, the ARM may be a much better tool as no variables are assigned as dependent or independent.

**Study Objectives**

The research team identified that the association of contributing factors leading to pedestrian crashes under different lighting conditions is still unexplored and there is scope for further research. The objective of this study is to offer a novel methodology, ARM, to be used to (a) identify the significant pattern of pedestrian crashes according to three different lighting conditions, (b) provide insights on how to choose appropriate countermeasures according to the identified crash pattern. The outcomes of this study would help researchers to a better understanding of pedestrian crash patterns in different lighting conditions and identify appropriate countermeasures.

**METHODS**
**Data Source**

Police-reported pedestrian crashes occurring within the period 2010-2019 were drawn from the Louisiana Department of Transportation and Development (DOTD) crash database. With the use of a matching criterion (crash number), the primary database was created by merging four data tables (pedestrian table, crash table, vehicle table, and DOTD table). The injury severity of a pedestrian in a crash is coded on an ABCDE scale (A = Fatal, B = Severe, C = Moderate, D = Complaint, E = No injury). To focus on only the evident injury, the research team decided to select only the fatal, severe, and moderate injury crashes (injury categories 'D' and 'E' do not exhibit any physical evidence of injury) (34). Henceforth, the term 'pedestrian crash' indicates fatal, severe, or moderate injury crashes only. Again, lighting condition in which a pedestrian crash occurs is coded on an ABDEFYZ scale (A = Daylight, B = Dark-no-streetlight, C = Dark-continuous streetlight, D = Streetlight at intersection only, E = Dusk, F = Dawn, Y = Unknown, Z = Other). To prepare the final database, only the fatal, severe, and moderate injury crashes occurring in the ABCD coded lighting condition were selected. Lighting conditions 'C' and 'D' were merged to get the condition 'dark-with-streetlight'. The final database comprises 8249 pedestrian injury crashes (fatal/severe/moderate) that occurred in three different lighting conditions i.e., daylight (3784, 45.8%), dark-with-streetlight (3042, 36.9%) and dark-no-streetlight (1423, 17.2%). The flowchart of database preparation is shown in **Figure 1.**

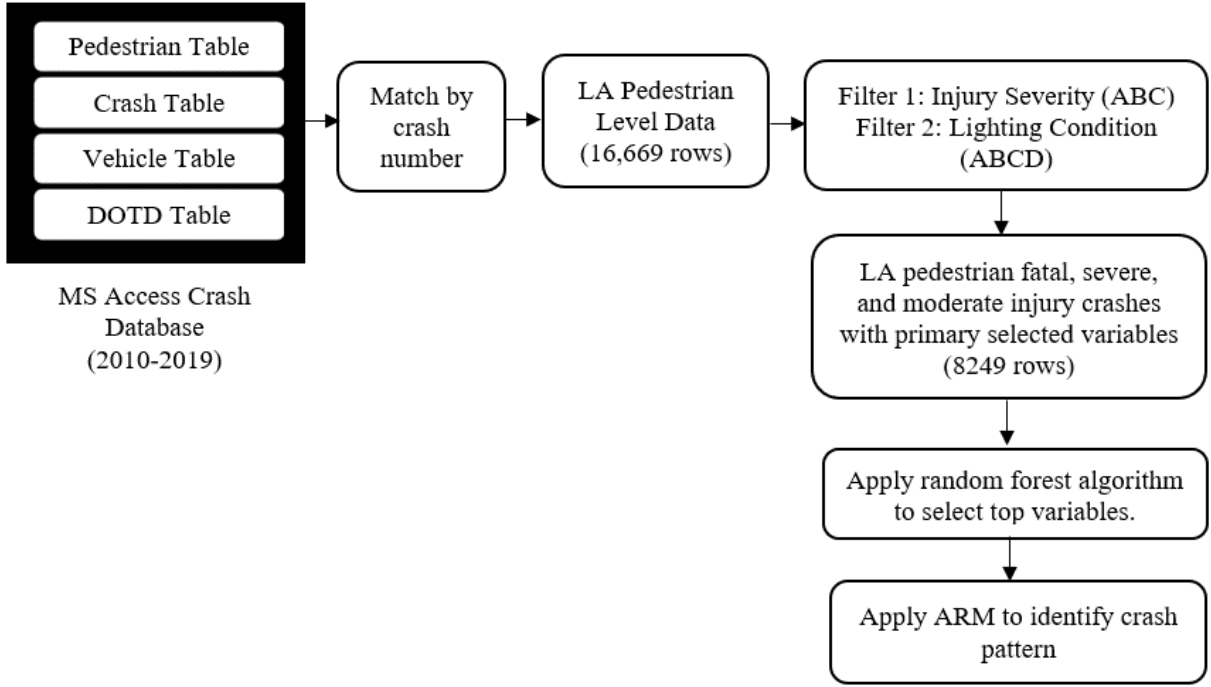

**Figure 1 Flow chart of database preparation**

**Methodology**

ARM, a rule-based machine learning method, helps in the identification of a group of items that appear together in a dataset. In 1993, Agarwal et al. proposed the 'Apriori' algorithm to uncover the pattern in supermarket transactions to determine which items are often brought together (35). After years of modification, ARM has become a more common approach for recognizing variable attribute patterns and interconnections in traffic safety-related research (36).

Let I = $\{i_1, i_2, \ldots \ldots \ldots, i_n\}$ is a set of n crash attributes called items (set of crash characteristics for each pedestrian crash record) and T = $\{t_1, t_2, \ldots \ldots \ldots, t_m\}$ is a database of pedestrian crash information such that each pedestrian crash record in T has a unique crash ID. Each $t_i \in I$ exemplifies each pedestrian crash record composed of a subset of items (chosen from I). An association rule can be written as Antecedent (A) → Consequent (K) where A and K are two separate subsets of all available items in the crash dataset. Three measures of significance are utilized in the generation of association rules. These are support, confidence, and lift, and the formulae for each are given below.

$$S(A) = \frac{\sigma(A)}{N} \tag{1}$$

$$S(K) = \frac{\sigma(K)}{N} \tag{2}$$

$$S(A \rightarrow K) = \frac{\sigma(A \cap K)}{N} \tag{3}$$

$$C (A \rightarrow K) = \frac{S (A \rightarrow K)}{S(A)} \tag{4}$$

$$L (A \rightarrow K) = \frac{S (A \rightarrow K)}{S(A). \, S(K)} \tag{5}$$

Where, N = number of crashes, $\sigma(A)$ = frequency of incidents with A, $\sigma(K)$ = frequency of incidents with K, $\sigma(A \cap K)$ = frequency of incidents with both A and K, $S(A \rightarrow K)$ = support of the association rule $(A \rightarrow K)$, $C (A \rightarrow K)$ = confidence of the association rule $(A \rightarrow K)$ and $L (A \rightarrow K)$ = lift of the association rule $(A \rightarrow K)$.

The support (S) implies how often antecedent (A) and consequent (K) of a particular rule appear together in the database, while the confidence (C) assesses the strength of the rule. The other measure lift (L) indicates the relationship between the antecedent-consequent co-occurrence frequency and expected frequency (37). If L > 1, it implies positive interdependence while L < 1 indicates negative interdependence between the antecedent and consequent. Lift value equal to 1 suggests that antecedent and consequent are independent and there is no correlation between them (38). The analysis was carried out with the help of the open-source program R version 4.0.1 and the R package 'arules' (39).

**RESULTS**
**Data Exploration**
    To analyze the association of contributing factors causing pedestrian crashes under different lighting conditions, 20 variables including 'lighting condition' were primarily selected (listed in **TABLE 1**). Note that percentages in the table may not add up to 100% due to rounding errors. Due to excessive skewness, several variables were not included which may influence the outcome of the association rules generation. For example, a majority of the pedestrian crashes occurred in straight alignment; thus, the roadway alignment variable was not included in the final database.

**TABLE 1 Overview of Pedestrian Crashes by Lighting Conditions**

| Type | Variable label | Variable categories | Daylight, N (%) | Dark-with streetlight, N (%) | Dark-no-streetlight, N (%) |
|---|---|---|---|---|---|
| Crash outcome | severity | fatal | 211 (5.6%) | 412 (13.5%) | 465 (32.7%) |
| | | severe | 507 (13.4%) | 658 (21.6%) | 257 (18.1%) |
| | | moderate | 3,066 (81%) | 1,972 (64.8%) | 701 (49.3%) |
| Pedestrian attributes | ped_action | crossing_intersection | 991 (26.2%) | 647 (21.3%) | 110 (7.7%) |
| | | crossing_midblock | 1,049 (27.7%) | 918 (30.2%) | 278 (19.5%) |
| | | walking_with_traffic | 221 (5.8%) | 344 (11.3%) | 346 (24.3%) |
| | | walking_against_traffic | 101 (2.7%) | 121 (4%) | 119 (8.4%) |
| | | other_inappropriate | 1,234 (32.6%) | 845 (27.8%) | 531 (37.3%) |
| | | unknown | 188 (5%) | 167 (5.5%) | 39 (2.7%) |
| | ped_alcohol_drug | yes | 156 (4.1%) | 524 (17.2%) | 370 (26%) |
| | | no | 2,665 (70.4%) | 1,396 (45.9%) | 568 (39.9%) |
| | | others | 963 (25.4%) | 1,122 (36.9%) | 485 (34.9%) |
| | ped_age | <15 | 882 (23.3%) | 228 (7.5%) | 58 (4.1%) |
| | | 15-24 | 649 (17.2%) | 632 (20.8%) | 297 (20.9%) |
| | | 25-40 | 765 (20.2%) | 892 (29.3%) | 515 (36.2%) |
| | | 41-64 | 1,058 (28%) | 1,070 (35.2%) | 460 (32.3%) |
| | | >64 | 365 (9.6%) | 144 (4.7%) | 68 (4.8%) |
| | | unknown | 65 (1.7%) | 76 (2.5%) | 25 (1.8%) |
| | ped_gender | male | 2,307 (61%) | 1,994 (65.5%) | 1,009 (70.9%) |
| | | female | 1,452 (38.4%) | 1,016 (33.4%) | 407 (28.6%) |
| | | unknown | 25 (0.7%) | 32 (1.1%) | 7 (0.5%) |
| | primary_factor | ped_action | 994 (26.3%) | 928 (30.5%) | 618 (43.4%) |
| | | ped_violation | 508 (13.4%) | 378 (12.4%) | 142 (10%) |
| | | ped_condition | 71 (1.9%) | 118 (3.9%) | 82 (5.8%) |
| | | prior_movement | 598 (15.8%) | 410 (13.5%) | 101 (7.1%) |
| | | other_factors | 1,613 (42.6%) | 1,208 (39.7%) | 480 (33.7%) |
| | ped_dark_cloth | yes | 955 (25.2%) | 1,214 (39.9%) | 700 (49.2%) |
| | | no | 2829 (74.8%) | 1828 (60.1%) | 723 (50.8%) |
| Driver and vehicle characteristics | driver_age | 15-24 | 567 (15%) | 456 (15%) | 242 (17%) |
| | | 25-34 | 714 (18.9%) | 585 (19.2%) | 305 (21.4%) |
| | | 35-44 | 580 (15.3%) | 404 (13.3%) | 218 (11.8%) |
| | | 45-54 | 500 (13.2%) | 343 (11.3%) | 168 (11.8%) |
| | | 55-64 | 421 (11.1%) | 269 (8.8%) | 127 (8.9%) |
| | | >64 | 433 (11.4%) | 203 (6.7%) | 98 (6.9%) |
| | | unknown | 569 (15%) | 782 (25.7%) | 265 (18.6%) |
| | driver_gender | male | 1,881 (49.7%) | 1,512 (49.7%) | 779 (54.7%) |
| | | female | 1,419 (37.5%) | 818 (26.9%) | 391 (27.5%) |
| | | unknown | 484 (12.8%) | 712 (23.4%) | 253 (17.8%) |
| | driver_condition | normal | 2,127 (56.2%) | 1,568 (51.5%) | 873 (61.3%) |
| | | inattentive_distracted | 763 (20.2%) | 310 (10.2%) | 130 (9.1%) |
| | | illness_fatigued_asleep | 28 (0.7%) | 8 (0.3%) | 8 (0.6%) |
| | | alcohol_drug | 96 (2.5%) | 226 (7.4%) | 92 (6.5%) |
| | | other_unknown | 770 (20.3%) | 930 (30.6%) | 320 (22.5%) |
| | violation_type | no_violations | 1,754 (46.4%) | 1,441 (47.4%) | 872 (61.3%) |
| | | careless_operation | 537 (14.2%) | 378 (12.4%) | 103 (7.2%) |
| | | failure_to_yield | 330 (8.7%) | 98 (3.2%) | 22 (1.5%) |
| | | others | 1,163 (30.7%) | 1,125 (37%) | 426 (29.9%) |
| | veh_type | passenger_car | 1,679 (44.4%) | 1,442 (47.4%) | 584 (41%) |
| | | van_suv | 933 (24.7%) | 649 (21.3%) | 278 (19.5%) |
| | | light_truck | 811 (21.4%) | 593 (19.5%) | 420 (29.5%) |
| | | others | 361 (9.5%) | 358 (11.8%) | 141 (9.9%) |
| Road and land use | location_type | business_industrial | 1,039 (27.5%) | 944 (31%) | 256 (18%) |
| | | business_mixed_residential | 1,193 (31.5%) | 1,211 (39.8%) | 361 (25.4%) |
| | | residential | 1,331 (35.2%) | 768 (25.2%) | 520 (36.5%) |
| | | open_country | 115 (3%) | 44 (1.4%) | 254 (17.8%) |
| | | other_locality | 106 (2.8%) | 75 (2.5%) | 32 (2.2%) |
| | road_type | one_way | 519 (13.7%) | 371 (12.2%) | 58 (4.1%) |
| | | two_no_separation (2-way with no physical separation) | 2,327 (61.5%) | 1,789 (58.8%) | 996 (70%) |
| | | two_separation (2-way with physical separation/barrier) | 864 (22.8%) | 844 (27.7%) | 366 (25.7%) |
| | | other_unknown | 74 (2%) | 38 (1.2%) | 3 (0.2%) |
| | highway_type | interstate | 109 (2.9%) | 116 (3.8%) | 148 (10.4%) |
| | | us_highway | 321 (8.5%) | 408 (13.4%) | 251 (17.6%) |
| | | state_highway | 604 (16%) | 648 (21.3%) | 556 (39.1%) |
| | | city_street | 2,202 (58.2%) | 1,554 (51.1%) | 264 (18.6%) |
| | | parish_road | 496 (13.1%) | 275 (9%) | 194 (13.9%) |
| | | others | 52 (1.4%) | 41 (1.3%) | 6 (0.4%) |
| | speed_limit (mph unit) | <30 | 1,439 (38%) | 754 (24.8%) | 162 (11.4%) |
| | | 30-35 | 1,201 (31.7%) | 1,035 (34%) | 212 (14.9%) |
| | | 40-45 | 558 (14.7%) | 771 (25.3%) | 387 (27.2%) |
| | | 50-55 | 172 (4.5%) | 178 (5.9%) | 424 (29.8%) |
| | | >55 | 109 (2.9%) | 103 (3.4%) | 173 (12.2%) |
| | | unknown | 305 (8.1%) | 201 (6.6%) | 65 (4.6%) |
| | intersection | yes | 1,555 (41.1%) | 1,261 (41.5%) | 257 (18.1%) |
| | | no | 2,229 (58.9%) | 1,781 (58.5%) | 1,166 (81.9%) |
| Other factors | day_of_week | weekday | 2,901 (76.7%) | 1,917 (63%) | 960 (67.5%) |
| | | weekend | 883 (23.3%) | 1,125 (37%) | 463 (32.5%) |
| | weather_condition | clear | 3,040 (80.3%) | 2,369 (77.9%) | 1,109 (77.9%) |
| | | cloudy | 532 (14.1%) | 365 (12%) | 185 (13%) |
| | | rain | 178 (4.7%) | 266 (8.7%) | 102 (7.2%) |
| | | fog_sleet_snow | 16 (0.4%) | 21 (0.7%) | 22 (1.5%) |
| | | other_unknown | 18 (0.5%) | 21 (0.7%) | 5 (0.4%) |

The table of descriptive statistics revealed a number of intriguing crash characteristics comparing three different lighting conditions. Several variable categories were overrepresented in the dark-no-streetlight condition compared to the other two lighting conditions. These were fatal pedestrian crashes (32.7% > 13.5% > 5.6%), walking with traffic (24.3% > 11.3% > 5.8%), pedestrian alcohol/drug presence (26% > 17.2% > 4.1%), pedestrian dark cloth presence (49.2% > 39.9% > 25.5%), interstate (10.4% > 3.8% > 2.9%), speed limit of more than 55 mph (12.2% > 3.4% > 2.9%). The comparison percentage inside the bracket indicates the sequence dark-no-streetlight > dark-with-streetlight > daylight. Pedestrian crashes involving '<15 years' age group were predominant in the daylight (23.3%) compared to the other two lighting conditions (dark-with-streetlight = 7.5%, dark-no-streetlight = 4.1%). Male pedestrians and drivers were more involved in crashes than the female counterparts, regardless of lighting conditions.

**Variable Selection by Random Forest**

From the primary list of variables, the Random Forest (RF) algorithm was applied to identify the significant variables with a high importance value. The RF approach is based on the bootstrap aggregation principle i.e., bagging principle (40). The study utilized Mean Decrease Accuracy (MDA) as a measure to identify the important variables based on a classification model (using lighting condition as the response variable and all other variables as explanatory). The variable importance plot shows how much accuracy is lost when each variable is removed from the model (41). A higher MDA value indicates that a variable is important for classifying, whereas a lower MDA value indicates the opposite. The following **Figure 2** shows the variable importance plot where each variable is shown on the y-axis and associated MDA on the x-axis. It is worth noting that, variables on the y-axis are ordered from most to least important (top to bottom).

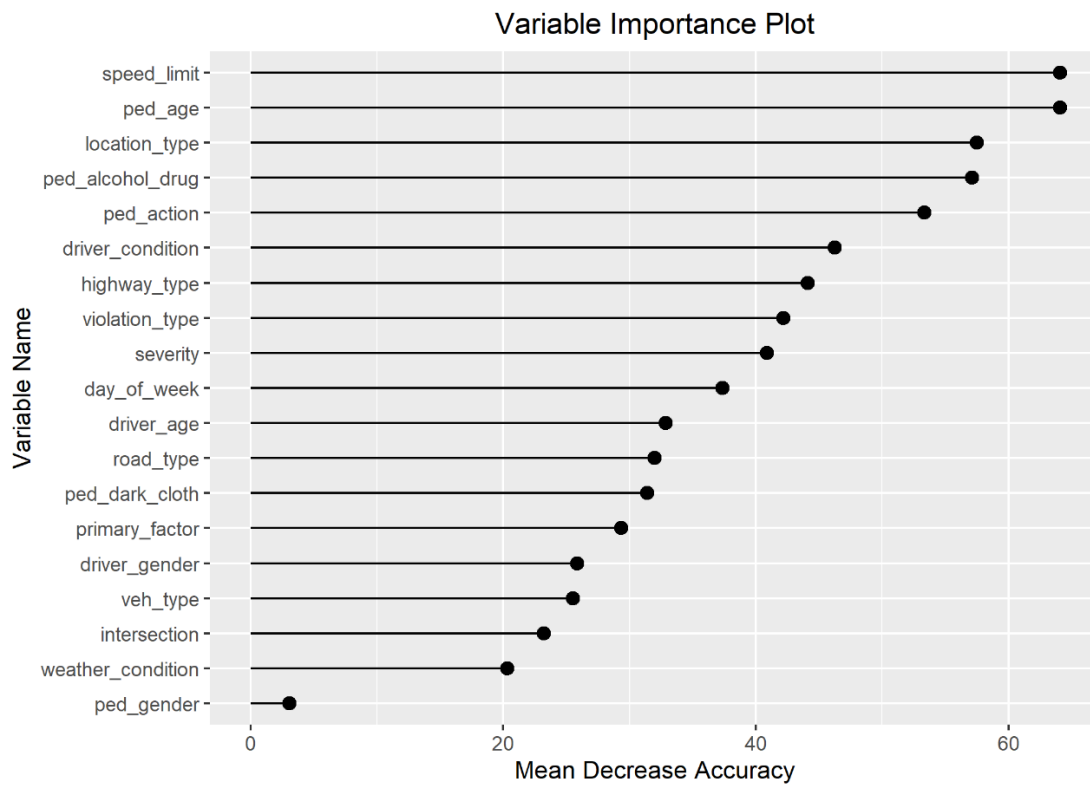

**Figure 2 Variable importance plot from random forest algorithm**

Since there is no certain standard for selecting the number of variables sorted according to MDA, a minimum MDA threshold value was defined above which the variables were selected. The top variable 'speed_limit' has an MDA value of 64.1. Fifty percentage of the top variable MDA value was selected as a minimum threshold value. In this way, the top thirteen variables were selected. The top thirteen variables selected for further analysis were:

- Speed limit
- Pedestrian age
- Location type
- Pedestrian alcohol/drug presence
- Pedestrian action
- Driver condition
- Highway type
- Violation type
- Severity
- Day of week
- Driver age
- Road type
- Pedestrian dark cloth

**Results of Association Rules Mining**

According to the primary analysis, the final dataset was comprised of 8,249 rows with 62 items (i.e., variable categories). The following absolute item frequency plot (**Figure 3**) shows numeric frequencies of each item independently in the entire dataset. The top five most frequently occurring items in the dataset were day of week = weekday, severity = moderate, pedestrian dark cloth = no, road type = two-way road with no physical separation, pedestrian alcohol/drug presence = no.

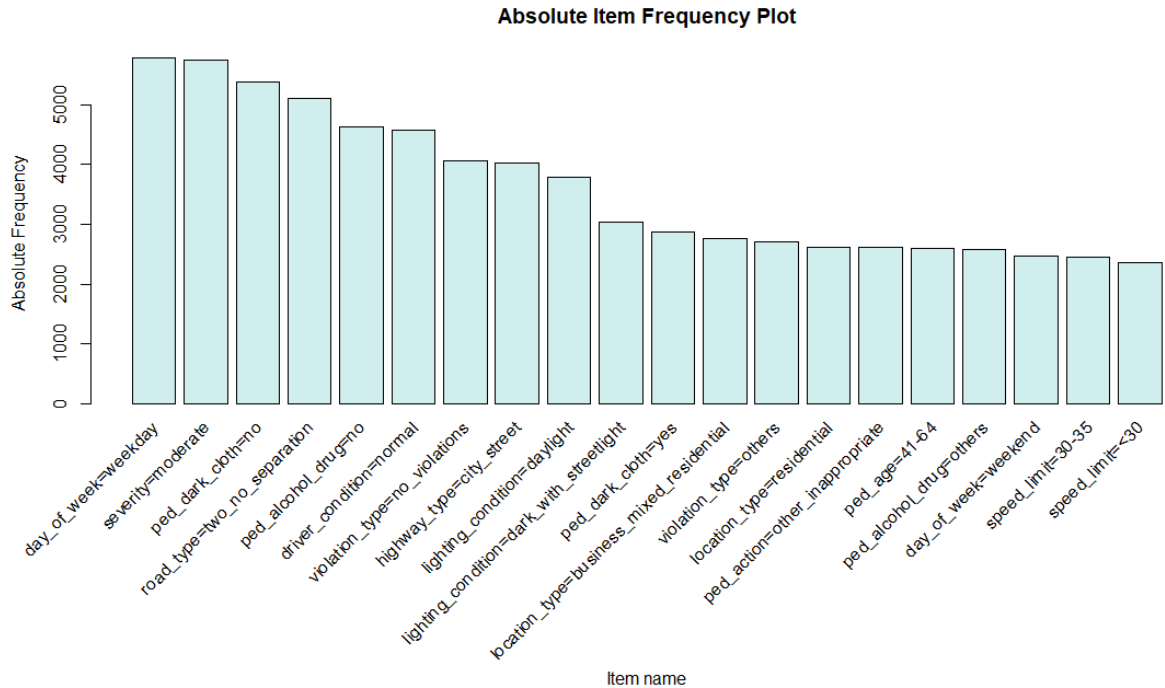

**Figure 3 Absolute item frequency plot**

The Apriori algorithm uses a 'bottom-up' technique to find association rules (42). To generate meaningful rules using ARM, it is critical to define an appropriate minimum level of support and confidence; otherwise, the algorithm could generate abundant rules. After a substantial number of trials and errors, the minimum value of support and confidence was selected. It might be argued that the values of these parameters (support and confidence) were subjective and determined on a case-by-case basis (21). In general, a high lift value suggests a stronger relationship between antecedents and consequents (43). Considering this point of view, the minimum threshold for lift value was chosen as 1.1. For ease of interpretation, the study was limited up to 4-itemsets rules. To identify and explain any given pattern associated with generated rules, a rule 'ID' was designed. Using the selected variables, ARM was applied to three separate scenarios (case 1: daylight, case 2: dark-with-streetlight, case 3: dark-no-streetlight).

*Case 1: Lighting Condition = Daylight*

The 'lighting_condition' variable was set to 'daylight' as right-hand-side (RHS) to mine the association rules for case 1. After several trials and errors, the minimum value of support and confidence was set at 0.1% and 50% respectively. Initially, the algorithm generated 6273 rules which contained a large number of redundant rules. After pruning redundant rules (44), 107 rules remained and were sorted according to descending order of lift value. **TABLE 2** lists the top 20 rules for this case.

**TABLE 2 Top 20 Association Rules for Pedestrian Crashes in Daylight Condition**

| ID | Antecedent (s) | S (%) | C (%) | L |
|----|----------------|-------|-------|---|
| R1 | ped_age = <15 | 10.69 | 75.51 | 1.65 |
| R2 | violation_type = failure_to_yield | 4.00 | 73.33 | 1.60 |
| R3 | ped_action = crossing_midblock, location_type = residential | 4.26 | 64.40 | 1.40 |
| R4 | road_type = other_unknown | 0.90 | 64.35 | 1.40 |
| R5 | driver_condition = illness_fatigued_asleep | 0.34 | 63.64 | 1.39 |
| R6 | location_type = other_locality, road_type = two_no_separation | 0.74 | 63.54 | 1.39 |
| R7 | driver_condition = inattentive_distracted | 9.25 | 63.42 | 1.38 |
| R8 | ped_age = >64 | 4.42 | 63.26 | 1.38 |
| R9 | ped_age =15-24, location_type = other_locality | 0.50 | 63.08 | 1.38 |
| R10 | ped_age = 41-64, day_of_week = weekend, location_type = other_locality | 0.12 | 62.50 | 1.36 |
| R11 | severity = severe, driver_age =15-24, driver_condition = other_unknown | 0.12 | 62.50 | 1.36 |
| R12 | highway_type = us_highway, driver_age =15-24, speed_limit = 30-35 | 0.28 | 62.16 | 1.36 |
| R13 | speed_limit = < 30 | 17.44 | 61.10 | 1.33 |
| R14 | location_type = residential, road_type = two_separation, violation_type = others | 0.57 | 60.26 | 1.31 |
| R15 | road_type = two_separation, driver_age =15-24, driver_condition = other_unknown | 0.18 | 60.00 | 1.31 |
| R16 | driver_condition = normal, violation_type = others | 3.66 | 59.80 | 1.30 |
| R17 | location_type = other_locality, driver_age = 15-24 | 0.42 | 59.32 | 1.29 |
| R18 | driver_age = >64 | 5.25 | 58.99 | 1.29 |
| R19 | ped_age = unknown, road_type = two_no_separation, driver_age = 25-34 | 0.12 | 58.82 | 1.28 |
| R20 | ped_age = unknown, location_type = business_industrial, violation_type = others | 0.12 | 58.82 | 1.28 |

First rule suggests that pedestrians under 15-years-old were highly associated with pedestrian crashes in the daylight condition (S = 10.69%, C = 75.51%, L = 1.65). The explanation of the first rule is: under 15-years-old pedestrians account for 10.69% of the crashes that occur in the daylight condition; out of all the '<15 years' old pedestrian crashes occurred in roadways, 75.51% took place in daylight condition; the proportion of under 15-years-old pedestrian crashes occurring on roadways during the daylight was 1.65 times the proportion of all '<15 years' old pedestrian crashes in the entire dataset. Cellphone distraction, listening to music while walking, and running out into the street to fetch a toy or ball are all plausible triggers for this type of daylight pedestrian crashes involving children and teenagers (45). Other pedestrian and driver age groups were found dominant in the generated rules (R8−R12, R15, R17, R18, R19). For illustration, pedestrians, and drivers aged 65 or higher were involved in crashes during the daylight (R8, R18). According to a previous study conducted by Zegeer et al., pedestrians 65 or older were overrepresented in crashes during the daylight hours (46). Another study found that elderly drivers (65 and up) were more involved in pedestrian deaths per unit of travel than other driver age groups (47). In general, an older person lacks vision, hearing, reaction time, and reduced ability to maintain attention which contributes to pedestrian crashes (48). Four other rules revealed the involvement of young drivers (15-24 years) in pedestrian crashes during daylight (R11, R12, R15, R17). Young drivers were overrepresented in these crashes possibly due to their lack of driving experience and risk-taking behavior.

Failure to yield to pedestrians resulted in crashes during the daylight (R2). Other driver-related factors such as distraction, inattention, and physical state (illness/fatigue/asleep) also contributed to pedestrian crashes at daylight (R5, R7). This result was supported by Stimpson et al. study which found that distracted drivers were responsible for the increasing share of pedestrian crashes (49). Different land-use patterns such as residential area (R3, R4), business/industrial area (R20) were also associated with pedestrian crashes in the daylight condition. According to previous studies, pedestrian collisions were more frequent in the residential and commercial areas due to increased pedestrian exposure during the daytime (50,51).

*Case 2: Lighting Condition = Dark-with-streetlight*
    To mine the association rules for case 2, the 'lighting_condition' variable was set to 'dark_with_streetlight' as a consequent. Following multiple rounds of trial and error, the minimal values of support (0.5%) and confidence (50%) were selected. The program initially produced 1,378 rules, many of which were repetitive. After pruning, 310 rules remained, which were sorted by lift value in descending order. The top 20 rules for this case are listed in **TABLE 3**.

**TABLE 3 Top 20 Association Rules for Pedestrian Crashes in Dark-with-streetlight Condition**

| ID | Antecedent | S (%) | C (%) | L |
|----|------------|-------|-------|---|
| R1 | ped_alcohol_drug = yes, highway_type = city_street | 2.55 | 68.85 | 1.87 |
| R2 | ped_alcohol_drug = yes, ped_action = crossing_intersection | 1.20 | 66.44 | 1.80 |
| R3 | ped_alcohol_drug = yes, speed_limit = 30-35 | 1.96 | 65.59 | 1.78 |
| R4 | driver_age = 25-34, speed_limit = 30-35, violation_type = careless_operation | 0.65 | 63.53 | 1.72 |
| R5 | ped_alcohol_drug = yes, speed_limit = < 30 | 1.15 | 62.91 | 1.71 |
| R6 | ped_alcohol_drug = yes, ped_action = crossing_midblock | 2.46 | 62.85 | 1.70 |
| R7 | ped_alcohol_drug = yes, location_type = business_mixed_residential | 2.89 | 62.47 | 1.69 |
| R8 | ped_alcohol_drug = yes, road_type = one_way | 0.58 | 62.34 | 1.69 |
| R9 | severity = fatal, ped_action = crossing_midblock, violation_type = others | 0.51 | 61.76 | 1.67 |
| R10 | ped_age = 25-40, driver_age = 25-34, violation_type = careless_operation | 0.52 | 61.43 | 1.67 |
| R11 | ped_alcohol_drug = yes, location_type = business_industrial | 1.90 | 60.62 | 1.64 |
| R12 | ped_alcohol_drug = yes, speed_limit = 40-45 | 2.18 | 60.20 | 1.63 |
| R13 | location_type = business_mixed_residential, speed_limit = 30-35, violation_type = careless_operation | 0.86 | 60.17 | 1.63 |
| R14 | ped_dark_cloth = yes, location_type = business_mixed_residential, violation_type = careless_operation | 0.67 | 59.78 | 1.62 |
| R15 | severity = severe, driver_age = unknown | 1.83 | 59.68 | 1.62 |
| R16 | day_of_week = weekend, highway_type = city_street, violation_type = careless_operation | 1.39 | 58.97 | 1.60 |
| R17 | ped_alcohol_drug = yes, driver_age = unknown | 1.36 | 58.95 | 1.60 |
| R18 | highway_type = state_highway, driver_condition = other_unknown, speed_limit = 30-35 | 0.57 | 58.75 | 1.59 |
| R19 | ped_alcohol_drug = yes, driver_age = 55-64 | 0.65 | 58.70 | 1.59 |
| R20 | ped_dark_cloth = yes, ped_action = crossing_midblock, location_type = business_industrial | 2.25 | 58.31 | 1.58 |

    The item {ped_alcohol_drug = yes} appears frequently (8 in top 20 rules) implying that alcohol/drug involved pedestrian crashes were often encountered in the dark-with-streetlight condition. This finding was supported by a prior study which found that alcohol/drug usage causes more nighttime pedestrian collisions (52). Pedestrian actions including crossing intersection (R2, L = 1.8) and crossing midblock (R6, L = 1.7) were also correlated with alcohol/drug-related pedestrian crashes. Impaired pedestrians are far more prone to conduct dangerous acts, such as crossing against the signal at an

intersection or crossing at road segments rather than using a designated crosswalk (53). 'Careless operation' was identified as a risk factor to pedestrian crashes for this lighting condition (R4, R10, R13, R14, R16) particularly among drivers in specific age groups (25-34 years).

Previous research suggested that pedestrians underestimate the significance of the nighttime conspicuity problem and therefore unintentionally place themselves at risk at night (54). Two of this study's association rules regarding pedestrian dark clothing have also identified this issue (R14, R20). Although different speed limits appear in the generated association rules, alcohol/drug-involved pedestrian crashes were identified both in low (< 30 mph) and high speed (40-45 mph) roadways in the dark-with-streetlight condition. The only fatal pedestrian crash was identified in rule 9 {severity = fatal, ped_action = crossing_midblock, violation_type = others} with a lift value of 1.67. This suggests that fatal pedestrian crashes occurred while crossing the street at midblock locations in the dark-with-streetlight condition. In general, pedestrians are less anticipated by drivers at midblock mainly because there might be no crosswalk markings, illumination enhancements, or other facilities (55). Businesses with mixed residential land use patterns were found to be a significant factor contributing to pedestrian crashes at dark-with-streetlight condition (R7, R13, R14). According to a prior study conducted by Wang and Kockelman, the mixing of residential and commercial land uses is linked to a higher probability of pedestrian crashes at various severity levels, and such a combination creates more potential conflicts between pedestrian and vehicle movements at night (56).

*Case 3: Lighting Condition = Dark-no-streetlight*

This is the most hazardous lighting condition for pedestrians to be involved in crashes. In this case, the 'lighting_condition' variable was set to 'dark_no_streetlight' as RHS. After numerous trials and errors, the minimum value of support and confidence was set at 0.4% and 55% correspondingly. Initially, the algorithm produced 510 rules which contained a large number of redundant rules. After trimming, 84 rules remained and were arranged according to descending order of lift value. **TABLE 4** lists the top 20 rules for this case.

**TABLE 4 Top 20 Association Rules for Pedestrian Crashes in Dark-no-streetlight Condition**

| ID | Antecedent | S (%) | C (%) | L |
|---|---|---|---|---|
| R1 | ped_action = walking_against_traffic, speed_limit = 50-55 | 0.53 | 73.33 | 4.25 |
| R2 | ped_action = walking_with_traffic, speed_limit = 50-55 | 1.47 | 72.02 | 4.18 |
| R3 | ped_alcohol_drug = yes, speed_limit = 50-55 | 1.81 | 71.29 | 4.13 |
| R4 | location_type = residential , highway_type = state_highway, driver_age = 25-34 | 0.73 | 67.42 | 3.91 |
| R5 | highway_type = us_highway, speed_limit = >55 | 0.52 | 67.19 | 3.89 |
| R6 | ped_dark_cloth = yes, speed_limit = 50-55 | 2.75 | 66.57 | 3.86 |
| R7 | ped_dark_cloth = yes, location_type = residential, highway_type = state_highway | 1.37 | 66.08 | 3.83 |
| R8 | ped_alcohol_drug = yes, ped_dark_cloth = yes, ped_action = walking_with_traffic | 0.68 | 65.88 | 3.82 |
| R9 | location_type = residential, speed_limit = 50-55 | 2.02 | 63.74 | 3.69 |
| R10 | ped_alcohol_drug = yes, speed_limit = >55 | 0.74 | 63.54 | 3.68 |
| R11 | severity = fatal, speed_limit = 50-55 | 2.19 | 63.51 | 3.68 |
| R12 | ped_age = 25-40, speed_limit = 50-55 | 1.93 | 63.35 | 3.67 |
| R13 | severity = fatal, ped_action = walking_against_traffic | 0.46 | 63.33 | 3.67 |
| R14 | driver_age = unknown, speed_limit = 50-55 | 0.73 | 63.16 | 3.66 |
| R15 | ped_alcohol_drug = yes, ped_action = walking_against_traffic, road_type = two_no_separation | 0.44 | 63.16 | 3.66 |

| R16 | ped_alcohol_drug = yes, severity = fatal, location_type = residential | 0.73 | 63.16 | 3.66 |
| R17 | ped_alcohol_drug = yes, severity = fatal, ped_action = other_inappropriate | 0.97 | 62.99 | 3.65 |
| R18 | ped_age = 25-40, location_type = residential, highway_type = state_highway | 1.08 | 62.68 | 3.63 |
| R19 | severity = fatal, speed_limit = >55 | 1.39 | 62.16 | 3.60 |
| R20 | road_type = two_no_separation, speed_limit = 50-55 | 4.17 | 62.09 | 3.60 |

The first two rules revealed an intriguing crash pattern relating to pedestrian actions and speed limits. Both these rules specified that pedestrians walking with or against the traffic on a road with a speed limit of 50-55 mph were involved in crashes in the dark without streetlight. Pai et al. discovered that pedestrians walking with the traffic were 1.26 times more likely to incur grievous injuries compared to those walking against the traffic (57). Some other noteworthy crash patterns linked to roadways with speed limits of 50-55 mph were pedestrian alcohol/drug involvement (R3, L = 4.13), pedestrians wearing dark cloth (R6, L = 3.86), residential area (R9, L = 3.69), pedestrian fatal injury (R11, L = 3.68), pedestrian age group 25 to 40 years (R12, L = 3.67) and two-way road without physical separation (R20, L = 3.60). It is important to note that just because a roadway has a speed limit of 50-55 mph does not mean that drivers adhere to it. According to previous observational studies, higher speed limits are strongly associated with higher vehicle speed and the chance of drivers exceeding the speed limit (58,59). Therefore, the absence of streetlights at night coupled with drivers exceeding the speed limit on high-speed roadways forms a lethal combination that contributes to pedestrian crashes.

The co-occurrence of inappropriate pedestrian actions and alcohol/drug involvement in rule 17 contributed to fatal pedestrian crashes in the dark-no-streetlight condition. This rule was expected because alcohol use by pedestrians impairs their judgment and cognitive ability leading to dangerous behaviors and getting involved in a crash (60). Rule 4 and 18 demonstrate that drivers of age group 25 to 34 years, as well as pedestrians of age group 25-40, are more likely to be involved in pedestrian crashes in residential areas on state highways during dark without streetlights.

## CONCLUSIONS

To explore the association of potential risk factors that influence pedestrian crashes under different lighting conditions, this study analyzed ten years of Louisiana pedestrian fatal and injury crash data with ARM. Some of the findings verify the common perceptions, while a few of others reveal quite interesting crash patterns in the daylight and dark (with and without streetlight) condition.

During the daylight, it is interesting to see that, the young and old driver age groups (15 to 24 years, and older than 64 years) showed a high likelihood of involving pedestrian crashes. Severe pedestrian crashes were associated with young drivers (15-24 years) in the daylight condition. This young driver age group was also associated with pedestrian crashes on roads with a speed limit between 30 and 35 mph, roadways with physical separation, as well as school zones. The lowest and highest pedestrian age groups (<15 years, >64 years) showed a high likelihood of involving crashes occurring during the day. It is also somewhat surprising to see that most of the daylight pedestrian crashes occurred on low-speed roadways (lower than 30 mph). Other crash risk factors at daylight condition were pedestrian action (crossing roadway at midblock), driver failure to yield, and driver physical condition such as illness, fatigue, and asleep.

At night with streetlight, fatal pedestrian crashes were associated with midblock crossing. Crashes with intoxicated pedestrians were found in most of the association rules in the dark-with-streetlight condition. Some of the risk factors associated with intoxicated pedestrian crashes were city street, pedestrians crossing at intersection and midblock location, speed limit, driver age (55-64 years), one-way street, and specific area type such as business/industrial area, business with mixed residential area. At night without streetlight, the presence of pedestrians in dark clothing and pedestrian intoxication was associated with a higher risk of crashes. Additionally, fatal pedestrian crashes were associated with high-speed limits (50 mph or higher) coupled with specific pedestrian action such as walking against the traffic and alcohol/drug intake in this lighting condition.

Improving pedestrian safety calls for problem-targeted countermeasures. Aiming at pedestrian crash risk factors identified by this study, the following countermeasures are recommended:

## Enforcement on intoxicated roadway users
The research team identified pedestrian alcohol/drug involvement as a serious safety issue in connection to pedestrian crashes at night with or without streetlights. The data analysis revealed a total of 136 underage drinking individuals as either driver (60%) or pedestrians (40%). As the Minimum Legal Drinking Age (MLDA) in the United States is 21 years (61), strong enforcement of this law would certainly eliminate or reduce such crashes, especially at night.

## Speed limit and speed control
Roadways with high-speed limits were associated with pedestrian fatalities in the dark without streetlight. Such locations with high pedestrian crashes may be considered to lower the speed limit. More awareness campaigns regarding the consequences of speeding should be carried out to shift the cultural mindset that does not see speeding as a serious problem. To reduce child (<15 years) pedestrian crashes during the daytime, greater enforcement of speed zones by installing speed control devices such as speed humps around schools and neighborhoods can play a vital role.

## Pedestrian warning devices
Several proven effective countermeasures can be used to alert drivers on pedestrian activities at night, such as encouraging pedestrians to wear fluorescent and retro-reflective materials, activated flashing LED warning sign systems, in-Road Warning Lights (IRWL), midblock pedestrian crossing signals such as HAWKS (High-Intensity Activated Crosswalk Signal), and RRFB (Rectangular Rapid Flash Beacon) (62,63). Another innovative method is to paint retroreflective signs at the entrance of a crosswalk to remind pedestrians to look in the right direction for oncoming traffic (64).

## Older pedestrian safety countermeasures
To improve safety for elderly pedestrians (>64 years), intervention measures focusing on education such as proper search behavior, enforcement, and regulation, roadway improvement (adjustment of pedestrian signal timing considering elderly people) activities can play a significant role (46).

## Built environment strategies
Separating pedestrians from motor vehicles by creating pedestrian safety zones, 4-way stops, advance stop lines, marked crosswalks, sidewalks, pedestrian overpasses, and fencing are some of the built environment strategies for improving pedestrian safety (45). Other viable solutions include traffic rerouting away from residential areas, pedestrian off-road pathways, and area-wide traffic calming (65).

## Technology
The current and future technology advancements can provide the best and inexpensive pedestrian crash countermeasures. The use of cell phone applications (based on Google Maps and GPS) to provide objective information (from one star least safe to five stars most safe) regarding pedestrian risks on specific walking routes may help to improve pedestrian safety (66). Other innovative technologies such as the installation of the pedestrian airbag in automobiles, which is designed to detect contact with an object having properties analogous to the human leg could reduce the number of lethal encounters between humans and automobiles (67).

It is impossible to significantly reduce pedestrian crashes by focusing on only a limited number of countermeasures. Therefore, another application of this research could be the development of strategies for breaking down associations to reduce the probability of pedestrian crashes. For illustration, the association rule of case 3: ID R8 was *{ped_alcohol_drug = yes, ped_dark_cloth = yes, ped_action = walking_with_traffic} => {lighting_condition = dark_no_streetlight}*. It indicates that alcohol/drug-involved pedestrians in dark clothing were involved in crashes while walking with the traffic at night without streetlight. Roadway lighting is already a proven safety countermeasure that can reduce pedestrian crashes by around 50% at night (68,69). However, simply installing streetlights does not ensure a reduction in these specific patterns of pedestrian crashes. Therefore, raising awareness to avoid drinking especially at night, and wearing retroreflective clothes in dark may reduce this kind of pedestrian crashes. It is recommended that safety officials go over other important association rules and seek strategies to break the association on a case-by-case basis. According to a recent report released by the United States Department of Transportation (USDOT), there is currently no consistent way to measure pedestrian crash risk exposure (70). The obtained significant rules in this study can serve as an 'exposure pattern' of pedestrian crashes according to the different lighting conditions.

The results of this research demonstrate the importance to use the holistic approach in pedestrian safety. However, this study is not without limitations. Determination of the most suitable value of support and confidence parameter is an issue that needs further investigation. The 'driver age' and 'pedestrian age' variables generated an 'unknown' category in some of the association rules because the original database had some 'unknown' category for about 19.43% of drivers and 2.01% of pedestrians. This is mainly due to the lack of input from the police report. Again, this suggests that any crash reporting system should be able to capture detailed information at the crash site to offer more accurate information on pedestrian crashes.

## ACKNOWLEDGMENTS

This research is sponsored by the Louisiana Transportation Research Center. The assistance and guidance provided by the paper review committee are appreciated. However, the authors are fully responsible for the results and discussions.

## AUTHOR CONTRIBUTIONS

The authors confirm contribution to the paper as follows: study conception and design: A. Hossain; data collection: A. Hossain; analysis and interpretation of results: A. Hossain, X. Sun, R. Thapa, J. Codjoe; draft manuscript preparation: A. Hossain, J. Codjoe. All the authors reviewed the results and approved the final version of the manuscript.